\documentclass{esannV2}
\usepackage[dvips]{graphicx}
\usepackage[latin1]{inputenc}
\usepackage{amssymb,amsmath,array}
\usepackage{algorithm}
\usepackage[noend]{algpseudocode}
\usepackage{array,multirow}
\usepackage{caption}
\begin{document}
\title{Evolutionary RL for Container Loading}
\author{S Saikia$^1$, R Verma$^1$, P Agarwal$^1$, G Shroff$^1$, L Vig$^1$ and A Srinivasan$^2$
%
\vspace{.3cm}\\
%
1- TCS Research - New Delhi
%
\vspace{.1cm}\\
2- Department of Computer Science, BITS-Pilani, Goa\\
}

\maketitle

\begin{abstract}
Loading the containers on the ship from a yard, is an important part of port operations. Finding the optimal sequence for the loading of containers, is known to be computationally hard and is an example of combinatorial optimization, which leads to the application of simple heuristics in practice. In this paper, we propose an approach which uses a mix of Evolutionary Strategies and Reinforcement Learning (RL) techniques to find an approximation of the optimal solution. The RL based agent uses the Policy Gradient method, an evolutionary reward strategy and a Pool of good (not-optimal) solutions to find the approximation. We find that the RL agent learns near-optimal solutions that outperforms the heuristic solutions. We also observe that the RL agent assisted with a pool generalizes better for unseen problems than an RL agent without a pool. We present our results on synthetic data as well as on subsets of real-world problems taken from container terminal. The results validate that our approach does comparatively better than the heuristics solutions available, and adapts to unseen problems better.
\end{abstract}

\section{Introduction}

The container shipping industry has been evolving lately in terms of the vessel size and  number of containers transported every day. Port operators want to automate the planning of small vessels and focus manual effort only for large ones. In this paper, we focus on the problem of generating the container loading schedule from yard to a ship, such that properties of each slot on the ship matches with that of the container to be filled. Container loading schedules that require the least rearrangement of containers in the yard(shuffles) are most desirable\cite{steenken2001stowage}.

Optimal container loading suffers from a non-polynomial increase of computation time with an increase in the number of containers. Therefore, heuristics-based solutions are commonly used. In the real world, external factors often lead to partial override of the planned loading sequence, leading to a worse loading schedule in practice. We model this problem as a Reinforcement Learning(RL) problem\cite{sutton1999reinforcement} that proposes the action to be taken in any current state, whereas heuristic solutions need to re-calculate the entire sequence in such situations.

Despite the recent achievements of deep Reinforcement Learning (deep RL) in complex gaming environments like Go \cite{silver2016mastering} and for increasing energy efficiency of cooling systems \cite{EvanandGao}, the use of RL for scheduling problems has been limited \cite{zhang1995reinforcement, hirashima2008intelligent}. To the best of our knowledge ours is the first attempt to solve container loading problem using deep RL, while it has been used for other problems of port operations such as container allocation in the yard \cite{hirashima2008intelligent} etc. Similar to RL, evolutionary strategies have also been used to solve similar problems \cite{tijjanicomparison}.

The key contributions of this paper are: 1)~Formulation of the container loading problem as an RL problem using a limited action space, 2)~A solution to container loading problem  using an evolutionary variant of RL, i.e., maintaining a pool of recently seen best solutions to train the agent using policy gradient, 3)~Usage of an adaptive reward function to train the RL agent, and 4)~Use of domain motivated intermediate rewards.

\section{Problem Description}
\label{prob_desc}

Slots on a ship should be loaded with containers in a specific order such as `sea-to-shore' to safeguard against ship tilting and an eventual topple event. Therefore, we consider a ship as a sequence of slots which need to be filled in order. Every slot has certain properties, such as `refrigerated' where containers with perishable items could be loaded. Combination of such properties of a slot is represented by a unique \textit{mask-id}. Containers to be loaded in these ship slots, also have an associated \textit{mask-id}, are stacked in a yard a priori. A container can be loaded into a slot of matching mask-id only.

When loading these containers in ship slots, if a container with the same mask-id as that of the next slot to fill, is not present in the yard on top of a stack, the yard operator will need to re-arrange the containers before picking the desired container. Such rearrangement is referred to as `shuffle'. The shuffle count of a container is taken as the number of containers stacked on top of the container. For example, in Fig. \ref{slot_yard}, the first slot of the ship (S0) can be filled using any of three matching containers from the yard (0-0-0, 0-0-3, or 6-0-5). Choosing 0-0-0 has no shuffle cost, choosing 0-0-3 will lead to a shuffle cost of 3 etc. The objective of container loading problem is to prescribe an order on the containers kept in the yard, to load them in ship slots in sequence, keeping the total shuffle count as low as possible.

\section{Approach and Solution}
\label{method}

\subsection{Reinforcement Learning (RL) Formulation}
\label{rl_formulation}

\textbf{\textit{Notations}}: We consider a sequence of ship slots $\{S_1, ..., S_N\}$ in which containers need to be loaded. The time step $t_i$ increments to the next time step $t_{i+1}$ after filling the current slot. Each slot is assigned one of the mask-id from the set $M~=~\{m_1, ..., m_k\}$. All containers $C_t=\{c_1, ..., c_n\}$, stacked in the yard at time $t$, are assigned unique mask-ids similar to ship slots. The function $\phi$ returns the mask-id of a slot or container, i.e., $\phi: S \cup C \rightarrow M$.Every container has a position identified by three dimensional representation, $P(c_j) = \{x, y, z\}$. Shuffle count $Q(c_j, t_i)$ of a container $c_j$ at time $t_i$, is defined as the number of containers placed above that container. Subset of $C_{t_i}$ with matching mask-ids of a slot $S_i$ is $C_{mt}(S_i, t_i) = \{c_j\}, \text{s.t. } \phi(S_i) = \phi(c_j), \forall c_j \in C_{t_i}$, where $1 \leq i \leq N, 0 \leq j \leq n$. The main objective of the container loading problem is to generate near-optimal loading sequences of containers, with minimum total shuffle count. 

\noindent\textit{\textbf{Action Space}}: The RL environment keeps picking the target slot to fill in a sequential manner. The action space becomes large with number of eligible containers as possible actions. 
In order to reduce this and to leverage heuristic optimization algorithms, our environment proposes a container $c_j \in C_{t_i}$ to be filled in the current slot, and passes it to the agent. The agent only needs to decide whether to agree with the environment's proposition or ask for an alternative container. Here, the environment proposes a container from the set $C_{mt}(S_i, t_i)$. If the agent chooses to agree with environment's proposition, mask-id of the container in the yard is marked with mask-id used for empty slots.

\begin{figure}
  \vspace{-20px}
  \centering
  \includegraphics{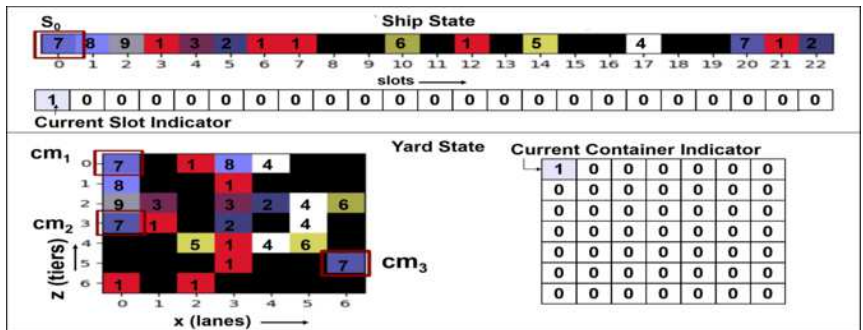}
   \vspace{-10px}
  \caption{An example of a state representation}
  \label{slot_yard}
\vspace{-20px}
\end{figure}

\noindent\textit{\textbf{State Space}}: The state space comprises of: i)~masking details of the ship slots, i.e., a vector of mask-ids; ii)~masking details of the containers in the yard, i.e., a vector (flattened) of mask-ids; iii)~target ship slot, i.e., one-hot vector with $1$ for current slot, and $0$ otherwise; and iv)~yard container proposed by the environment, i.e., a one-hot vector. Fig. \ref{slot_yard} depicts an example of state representation, here a different color is used for every mask-id, and an integer mask-id is indicated for every container and slot. The current slot to fill $S_0$ and the set of the matching containers $C_{mt}(S_0, t_0)$ are highlighted with red rectangles. 

\noindent\textit{\textbf{Reward}}: We have used both final $R_f$ and intermediate rewards $R_i$ to facilitate learning. $R_f = 1$ if the total shuffle count of the solution is smaller than a threshold $\theta$, and -1 otherwise. We use an evolving $\theta$, i.e., the value of $\theta$ changes as the RL agent learns. For the best solution encountered so far for a problem, $R_f = 2$. The intermediate reward $R_i$, for picking a container $c_j$ is modeled based on domain specific rule derived using i)~shuffle count $Q(c_j, t_i)$ of container $c_j$, and ii)~number of good containers uncovered $u(c_j)$. A container is referred to as a good container if it can be used to fill an upcoming slot, and it is said to be uncovered if current action reduces its shuffle count in future. We model intermediate reward as $R_i = 0.1$, if $Q(c_j, t_i) = 0$, or $R_i = 0.05 + 0.1 \times u(c_j)/6$, if $Q(c_j, t_i) \in \{1, 2\}$, or $R_i = 0.1 \times u(c_j)/6$, if $Q(c_j, t_i) = 3$, and zero otherwise. It ensures instantaneous feedback during training. 

\subsection{Training Procedure}
We train the agent using policy gradient, which uses a deep feed forward neural network (policy network). Generating a full sequence of containers to be loaded for a problem is referred to as an \textit{episode}. The agent uses the policy network to decide whether it should agree with environment's proposition such that it leads to minimum total shuffle cost at the end of the episode. The episode terminates if all slots in the ship are filled or if there are no containers left in the yard. 

Running of an episode generates many state-action pairs and their corresponding discounted rewards, $v_i = R_i + R_f \times \gamma^{t}$. Here $\gamma$ is the discount factor and $t$ is the sequence number of the action (taken in reverse order of the episode). If the episode is positively rewarded at the end we push all state-action pairs and their rewards into a priority queue(based on their reward), hereafter referred to as \textit{pool}. We train the policy network at the end of every episode, picking top-K state action pairs from the pool. 

The training procedure involves many iterations of such training, here an iteration includes $E$ episodes of every problem configuration of training data. We adopt an evolutionary rewarding strategy for training the RL agent, after $k$ ($ k <E $) episodes of each problem, we change the value of $\theta$ to average shuffle count of $k$ episodes, if it is smaller. This results in an adaptive reward function.

Our intended loss for policy gradient is $\mathcal{L} = \sum_{i} A_i \log p(a_i|s_i)$, where $a_i$ is the action we sampled when the state $s_i$ was observed and $A_i$ is the advantage. We want to increase the log probability for actions that worked and decrease it for those that did not work, so we use the discounted reward of a state-action pair $v_i$ as the value of $A_i$. Since this loss function is identical to categorical cross entropy loss function, we use it for training the neural network.

\subsection{Fusion of Evolutionary Strategy and Policy Gradient}
When using pure policy gradient based training we find it hard to train the agent because a)~it does not generalize, i.e., after learning to solve some more problems, it would forget to solve the problem which it could solve earlier b)~finding the right reward function to train on different problems is a challenge, since it always agrees or always disagrees with environment's proposition.

Our implementation of RL is a fusion of policy gradient and evolutionary strategies (ES). ES uses a population of solutions to train a model. This population of solutions is maintained based on an evolving threshold on an objective function. The evolving threshold finally leads to a near optimal solution. For each problem, we maintain a distinct pool of solutions to be used by the RL agent to learn from using the policy gradient technique. This pool stores the state-action pairs along with their rewards. When training the agent at the end of an episode, we pick top-K episodes for the problem from the pool.

\section{Experiments and Results}
\label{results}

To evaluate our approach as described in Section \ref{method}, we generated synthetic data inspired from the real data, and used it to train the RL agent as well as to test the accuracy of our approach. The data contains 23 ship slots, of which 15 need to be filled from a block of 49 containers organized as 7 stacks of containers each having 7 containers in it. We use the real data to only test the RL system.

\textit{\textbf{Baselines Approaches}}: Since the container loading problem can be modelled as that of combinatorial explosion, it can be solved using backtrack search. However, it takes in-ordinate amount of time to solve this problem with large number of ship slots, i.e., for more than 24 slots, backtrack search takes many days to converge. An approach where we pick a container considering $k$ subsequent ship slots, is termed as k-step lookahead policy. In this paper, we present a comparison of our approach with the 0-step lookahead, 1-step lookahead and optimal result obtained using backtrack search. On a test set of 900 problems, 0-step policy gives optimal solution for 333 problems, 1-step policy gives for 522 problems and random policy for only 5 problems. 

\textit{\textbf{Environment}}: We design an OpenAI Gym (https://gym.openai.com) environment for the container loading problem, described in Section \ref{method}. The environment proposes one of the matching containers for a target slot to the agent randomly. If there are more than one matching containers present in a stack, then the environment would propose the top-most container only. 

\begin{figure}[h!]
 \vspace{-10px}
  \centering
  \includegraphics{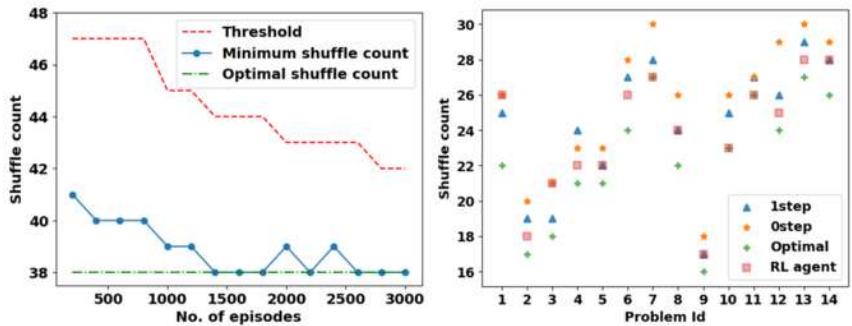}
  \vspace{-10px}
  \caption{(a) Evolving threshold during training for a problem (b)Minimum shuffle counts by various policies for 14 problems}
  \vspace{-20px}
  \label{RLplots}
\end{figure}

\textit{\textbf{Hyper-parameter Settings}}: The policy network of the agent takes state space consisting $144 (=7\times 7 \times 2 + 23 \times 2)$ input units. The neural network has three hidden layers of size $128\times64\times32$. During training, we update reward threshold $\theta_{new} = \min(\theta_{curr}, \textit{\text{avg\_shuffle\_ct}})$, here $\textit{\text{avg\_shuffle\_ct}}$ is average shuffle count of last $20$ episodes. Thus, the network indulges in self-play, comparing itself with its previous performances, and adapting the threshold to compute the reward. At each timestep, $top\ 50$ good examples from the pool are used to train the agent.
Training dataset comprises of 80 problems from the synthetic dataset. The training algorithm runs over a series of these 80 problems for four iterations, playing $N=200$ episodes per problem in each iteration. Evaluation is conducted under these four experimental settings: delayed reward without pool (DRWP), delayed reward with pool (DRP), intermediate and delayed reward without pool (IRWP) and intermediate and delayed reward with pool (IRP). The results are presented in Table \ref{Tab:tab1}.
\begin{table*}
\resizebox{\textwidth}{!}{%
  \begin{tabular}{@{\extracolsep{\fill}}|l|c|c|c|c|c|c|}
    \hline
    Experiment & $<$0 step & $\leq$0 step & $<$1 step & $\leq$1 step & $>$Optimal & $=$Optimal \\
    \hline
    DRWP  &4.5&86.3&5.6&61.9&64.5&35.4\\
    \hline
     DRP  &18.9 &89.6&8.7&62.1&64.2&35.7\\
     \hline
    IRWP  &5.3&87.8&5.2&66.9&61.8&38.2\\
    \hline
    \textbf{IRP}  & \textbf{26.7} & \textbf{97.8}  & \textbf{12.5} & \textbf{74.4}  & \textbf{52.6} & \textbf{47.3}\\
    \hline
  \end{tabular}}
  \vspace{-10px}
  \caption{Performance statistics (in \%) of RL agent on synthetic data.}\label{Tab:tab1}
  \vspace{-20px}
\end{table*}
The intermediate reward is computed after every timestep during training, whereas, the delayed reward is received by the RL agent at the end of each episode. The results show that the agent performs better with the proposed intermediate reward. Also, using a pool of good examples during training helps the agent to generalize better on unseen problems, as evident from the results. \\
The plots in Fig. \ref{RLplots} show the evolutionary behaviour of the RL agent as many episodes run on a problem while performing self-play. It also shows how random, heuristic, optimal and RL policies perform on a subset of training data. The threshold in the first plot keeps dropping till the minimum shuffle count reaches the optimal. We further evaluate the IRP model on real-world dataset of 11 problems. It performs at par with the 0-step policy on all 11 problems and better on 5 problems. However, it is not able to match the results of the 1-step policy. This is because the synthetic data used for training is different from the real-world data as there is not much variation in mask-ids of containers in the yard and the slots, in the latter. Also, in practice the yard is much larger and we are yet to generalize our approach to solve such large problems.

\section{Conclusion}
\label{conclusion}
We described and formulated container loading problem, as a reinforcement learning problem, which is often needed at the shipping ports. We introduced evolutionary Reinforcement learning, which uses a pool of good training examples, as normally performed in evolutionary learning. We have also demonstrated that for the container loading problem evolutionary reinforcement learning performs better than pure policy gradient learning. We also demonstrate that domain based intermediate reward helps the RL agent in learning.

\begin{footnotesize}
\bibliographystyle{unsrt}
\bibliography{rl_paper}

\end{footnotesize}
\end{document}